\documentclass[11pt,a4paper]{article}
\usepackage{times,latexsym}
\usepackage{url}
\usepackage[T1]{fontenc}

\usepackage[]{acl}

\usepackage{xspace,mfirstuc,tabulary}

\usepackage{xspace}
\usepackage{amsmath}
\usepackage{amssymb}
\usepackage{amsfonts}
\usepackage{microtype}
\usepackage{bm}
\usepackage[ruled,noend]{algorithm2e}
\usepackage{setspace}
\usepackage{graphicx}
\usepackage{enumitem}

\SetKwInput{KwInput}{Input}
\SetKwInput{KwOutput}{Output}
\SetEndCharOfAlgoLine{}

\newcommand{\modelname}{\texttt{Prix-LM}\xspace}

\newcommand{\stitle}[1]{\vspace{1.2mm}\noindent{\bf #1}}
\newcommand{\rtitle}[1]{\noindent{\bf #1}}

\newcommand{\lpresults}[9]{ #1 & #2 & #9 & #4 & #5 & #3 & #7 & #6 & #8 & \foocontinued}
\newcommand\foocontinued[2]{%
    #1 & #2  
}

\usepackage{booktabs}
\usepackage{multirow}
\usepackage{colortbl}

\usepackage{cleveref}
\crefformat{section}{\S#2#1#3}
\crefformat{subsection}{\S#2#1#3}
\crefformat{subsubsection}{\S#2#1#3}
\crefrangeformat{section}{\S\S#3#1#4 to~#5#2#6}
\crefmultiformat{section}{\S\S#2#1#3}{ and~#2#1#3}{, #2#1#3}{ and~#2#1#3}
\usepackage{refstyle}
\Crefformat{figure}{#2Fig.~#1#3}
\Crefmultiformat{figure}{Figs.~#2#1#3}{ and~#2#1#3}{, #2#1#3}{ and~#2#1#3}
\Crefformat{table}{#2Tab.~#1#3}
\Crefmultiformat{table}{Tabs.~#2#1#3}{ and~#2#1#3}{, #2#1#3}{ and~#2#1#3}
\Crefformat{appendix}{Appx.~\S#2#1#3}
\crefformat{algorithm}{Alg.~#2#1#3}

\usepackage{xspace}
\newcommand{\en}{{\textsc{en}}\xspace}
\newcommand{\fin}{{\textsc{fi}}\xspace}
\newcommand{\de}{{\textsc{de}}\xspace}
\newcommand{\es}{{\textsc{es}}\xspace}
\newcommand{\zh}{{\textsc{zh}}\xspace}
\newcommand{\tr}{{\textsc{tr}}\xspace}
\newcommand{\ru}{{\textsc{ru}}\xspace}
\newcommand{\fr}{{\textsc{fr}}\xspace}

\newcommand{\ita}{{\textsc{it}}\xspace}

\newcommand{\et}{{\textsc{et}}\xspace}

\newcommand{\hu}{{\textsc{hu}}\xspace}
\newcommand{\ja}{{\textsc{ja}}\xspace}
\newcommand{\ko}{{\textsc{ko}}\xspace}

\newcommand{\te}{{\textsc{te}}\xspace}
\newcommand{\lo}{{\textsc{lo}}\xspace}
\newcommand{\mr}{{\textsc{mr}}\xspace}

\newcommand{\mask}{\textsc{[Mask]}}
\newcommand{\rel}[1]{\verb~#1~}
\newcommand{\ent}[1]{\textsc{#1}}
\usepackage{txfonts}

\usepackage{tcolorbox}

\usepackage{todonotes}
\makeatletter
\newcommand*\iftodonotes{\if@todonotes@disabled\expandafter\@secondoftwo\else\expandafter\@firstoftwo\fi}
\makeatother

\title{\modelname: Pretraining for Multilingual Knowledge Base Construction}

\author{Wenxuan Zhou$^{1*}$, Fangyu Liu$^2$\thanks{\;\;Indicating equal contribution.}\;, Ivan Vuli\'{c}$^2$, Nigel Collier$^2$, Muhao Chen$^1$\\
$^1$LUKA Lab, University of Southern California, USA\\
$^2$Language Technology Lab, TAL, University of Cambridge, UK\\
\texttt{\{zhouwenx,muhaoche\}@usc.edu}\; \texttt{\{fl399,iv250,nhc30\}@cam.ac.uk}
}

\date{}

\begin{document}
\maketitle

\begin{abstract}
Knowledge bases~(KBs) contain plenty of structured world and commonsense knowledge. As such, they often complement distributional text-based information and facilitate various downstream tasks. Since their manual construction is resource- and time-intensive, recent efforts have tried leveraging large pretrained language models~(PLMs) to generate additional monolingual knowledge facts for KBs. However, such methods have not been attempted for building and enriching multilingual KBs. Besides wider application, such multilingual KBs can provide richer combined knowledge than monolingual (e.g., English) KBs. Knowledge expressed in different languages may be complementary and unequally distributed: this implies that the knowledge available in high-resource languages can be transferred to low-resource ones. To achieve this, it is crucial to represent multilingual knowledge in a shared/unified space.
To this end, we propose a unified representation model, \modelname \includegraphics[height=1em]{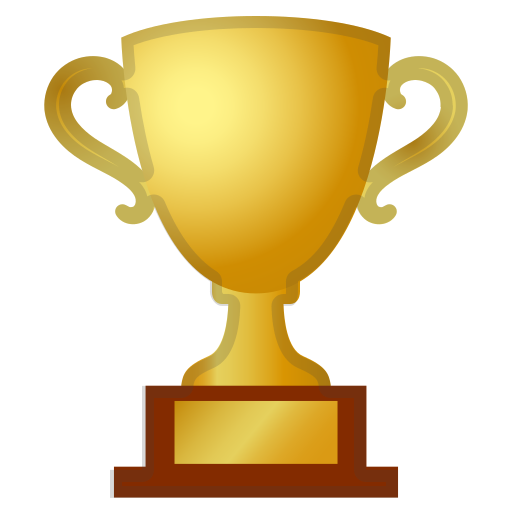}, for multilingual KB construction and completion. We leverage two types of knowledge, \textit{monolingual triples} and \textit{cross-lingual links}, extracted from existing multilingual KBs, and tune a multilingual language encoder XLM-R via a causal language modeling objective. \modelname integrates useful multilingual and KB-based factual knowledge into a single model. 
Experiments on standard entity-related tasks, such as link prediction in multiple languages, cross-lingual entity linking and bilingual lexicon induction, demonstrate its effectiveness, with gains reported over strong task-specialised baselines.
\end{abstract}

\section{Introduction}

\begin{figure}[t]
    \centering
    \includegraphics[width=0.95\linewidth]{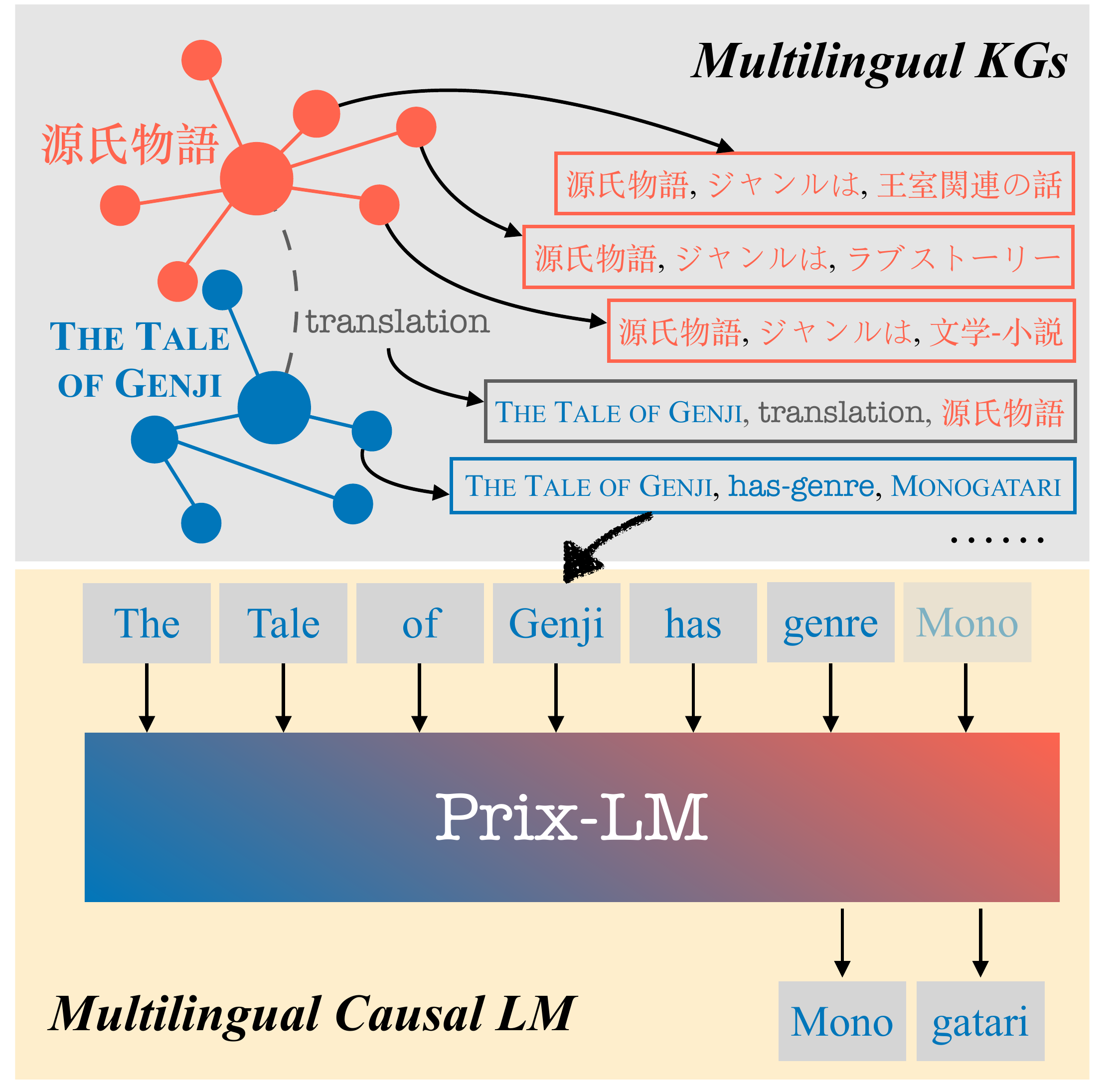}
    \caption{An illustration of the main idea supporting \modelname: it infuses complementary multilingual knowledge from KGs into a multilingual causal LM; e.g., Japanese KG stores more comprehensive genre information of \ent{The Tale of Genji} than KGs in other languages. Through cross-lingual links (translations), such knowledge is then propagated across languages.}
    \label{fig:front_fig}
\end{figure}



Multilingual knowledge bases (KBs), such as DBPedia~\cite{lehmann2015dbpedia}, Wikidata~\cite{wikidata14}, and YAGO~\cite{yago07}, 
provide structured knowledge expressed in multiple languages.
Those KBs are modeled as knowledge graphs (KGs) that possess two types of  knowledge: \emph{monolingual triples} which describe relations of entities, and \emph{cross-lingual links} which match entities across languages.
The knowledge stored in such KGs facilitates various downstream applications such as question answering~\cite{dai-etal-2016-cfo,bauer-etal-2018-commonsense, wang-etal-2021-k}, recommendation~\cite{zhang2016collaborative,wang2018dkn,wang2021learning}, and dialogue systems~\cite{madotto-etal-2018-mem2seq,liu-etal-2019-knowledge,yang-etal-2020-graphdialog}.

Manually constructing large-scale knowledge bases has been labor-intensive and expensive \cite{paulheim2018much}, leading to a surge of interest in automatic knowledge base construction~\cite{ji2021survey}.
Recent research~\cite[\textit{inter alia}]{bosselut-etal-2019-comet,Yao2019KGBERTBF,wang2020language} proposes to generate structured knowledge using pretrained language models (PLMs; \citealt{devlin-etal-2019-bert}), where missing elements in KB facts (i.e., triples) can be completed (i.e., filled in) by the PLM.

While these methods arguably perform well for English, such automatic KB construction has not yet been tried for multilingual KBs -- improving the knowledge in multilingual KBs would have a positive impact on applications in other languages beyond English. Moreover, KBs in multiple languages 
may possess complementary knowledge, and knowledge bases in low-resource languages often suffer severely from missing entities and facts. This issue could be mitigated by propagating knowledge from multiple well-populated high-resource languages' KBs (e.g., English and French KBs) to the KBs of low-resource languages, this way `collectively' improving the content stored in the full multilingual KB.\footnote{This intuition is illustrated by the example in Figure~\ref{fig:front_fig}. Consider the prediction of facts (e.g., \textit{genre}) about the oldest Japanese novel \ent{The Tale of Genji}. English DBpedia records its genre only as \emph{Monogatari} (story), whereas complementary knowledge can be propagated from the Japanese KB, which provides finer-grained genre information, including \textit{Love Story}, \textit{Royal Family Related Story}, and \textit{Monogatari}.}

However, training LMs to capture structural knowledge independently for each language will fall short of utilizing complementary and transferable knowledge available in other languages.
Therefore, a unified representation model is required, which can capture, propagate and enrich knowledge in multilingual KBs. In this work, we thus propose to train a language model for constructing multilingual KBs.
Starting from XLM-R~\cite{conneau-etal-2020-unsupervised} as our base model, we then pretrain it on the multilingual DBpedia, which stores both monolingual triples and cross-lingual links (see Figure~\ref{fig:front_fig}).
We transform both types of knowledge into sequences of tokens and pretrain the language model with a causal LM objective on such transformed sequences. The monolingual triples infuse structured knowledge into the language model, while the cross-lingual links help align knowledge between different languages.
This way, the proposed model \modelname \includegraphics[height=1em]{pics/1f3c6.png} (\underline{Pr}e-trained Knowledge-\underline{i}ncorporated \underline{Cross}-lingual \underline{L}anguage \underline{M}odel) is capable of mapping knowledge of different languages into a unified/shared space.

We evaluate our model on four different tasks essential for automatic KB construction, covering both high-resource and low-resource languages: link prediction, cross-lingual entity linking, bilingual lexicon induction, and prompt-based LM knowledge probing.
The main results across all tasks indicate that \modelname brings consistent and substantial gains over various state-of-the-art methods, demonstrating its effectiveness. 

\section{\modelname}
We now describe \modelname, first outlining the data structure and pretraining task, and then describing its pretraining procedure in full (\Cref{ssec:PLM}), and efficient inference approaches with \modelname (\Cref{ssec:inference}).

\stitle{Pretraining Task.}
\label{ssec:pretrain_task}
We rely on multilingual DBpedia, but note that \modelname is also applicable to other KBs. DBpedia contains two types of structured knowledge: monolingual knowledge triples, and cross-lingual links between entities. The monolingual triples represent (relational) facts expressed in a structured manner. Each triple is denoted as $\{e_1, r, e_2\}$: the elements of a triple are identified as the subject entity $e_1$, relation (or predicate) $r$, and object entity $e_2$, respectively (see also Figure~\ref{fig:front_fig} for examples).
For instance, the fact ``\textit{The capital of England is London}'' can be represented as $\{\ent{England}, \rel{capital}, \ent{London}\}$.
The cross-lingual links, denoted as $\{e_a, e_b\}$, represent the correspondence of `meaning-identical' entities $e_a$ and $e_b$ in two different languages: e.g., the English entity \ent{London} is mapped to \ent{Londres} in Spanish.

We treat both types of knowledge using the same input format $\{s, p, o\}$, where $s=e_1, p=r, o=e_2$ for monolingual knowledge triples, and $s=e_a, p=\textit{null}, o=e_b$ for cross-lingual entity links.
The pretraining task is then generating $o$ given $s$ and $p$. This objective is consistent with the link prediction task and also benefits other entity-related downstream tasks, as empirically validated later. 


\subsection{Pretraining Language Models}\label{ssec:PLM}
\modelname is initialized by a multilingual PLM such as XLM-R~\cite{conneau-etal-2020-unsupervised}:
starting from XLM-R's pretrained weights, we train on the structured knowledge from a multilingual KB.

\stitle{Input Representation.}
We represent knowledge from the KB as sequences of tokens.
In particular, given some knowledge fact  $\{s, p, o\}$, where each element is the surface name of an entity or a relation, we tokenize\footnote{XLM-R's dedicated multilingual tokenizer is used to processes entity and relation names in each language.} the elements to sequences of subtokens $X_s$, $X_p$, and $X_o$. 
We treat each element in the knowledge fact as a different text segment and concatenate them to form a single sequence.
We further introduce special tokens to represent different types of knowledge:

\vspace{0.8mm}
\noindent \textit{(1) Monolingual Triples.}
We use special tokens to indicate the role of each element in the triple, which converts the sequence to the following format:
\begin{tcolorbox}
\small
        \texttt{<s>} \texttt{[S]}\boldsymbol{$X_s$} \texttt{</s>}
        \texttt{</s>} \texttt{[P]}\boldsymbol{$X_p$} \texttt{</s>}
        \texttt{</s>} \texttt{[O]}\boldsymbol{$X_o$} \texttt{[EOS]}\texttt{</s>}.
\end{tcolorbox}
    \texttt{<s>} is the special token denoting beginning of sequence; \texttt{</s>} is the separator token, both adopted from XLM-R. Additional special tokens \texttt{[S]}, \texttt{[P]} and \texttt{[O]} denote the respective roles of subject, predicate, and object of the input knowledge fact. \texttt{[EOS]} is the 
    end-of-sequence token.
    
    \vspace{0.6mm}
    \noindent \textit{(2) Cross-Lingual Links.}
    As the same surface form of an entity can be associated with more than language, we use special language tokens to indicate the actual language of each entity. These extra tokens can also be interpreted as the relation between entities. The processed sequence obtains the following format: 
\begin{tcolorbox}
\small
    \texttt{<s>} \texttt{[S]}\boldsymbol{$X_s$} \texttt{</s>}
    \texttt{</s>} \texttt{[P]}\texttt{[\emph{S-LAN}]}\texttt{[\emph{O-LAN}]} \texttt{</s>}
    \texttt{</s>} \texttt{[O]}\boldsymbol{$X_o$} \texttt{[EOS]}\texttt{</s>}.
\end{tcolorbox}
    \texttt{<s>} and \texttt{</s>} are the same as for monolingual triples.  
    \texttt{[\emph{S-LAN}]} and \texttt{[\emph{O-LAN}]} denote two placeholders for language tokens, where they get replaced by the two-character ISO 639-1 codes of the source and target language, respectively. For example, if the cross-lingual connects an English entity \ent{London} to a Spanish entity \ent{Londres}, the two language tokens \texttt{[\emph{EN}]}\texttt{[\emph{ES}]} will be appended to the token \texttt{[P]}.
The new special tokens are randomly initialized, and optimized during training. 
The original special tokens are are kept and also optimized.

\stitle{Training Objective.}
The main training objective of \modelname is to perform completion of both monolingual knowledge triples and cross-lingual entity links (see \S\ref{ssec:pretrain_task}). In particular, given $X_s$ and $X_p$, the model must predict 1) $X_o$ from monolingual triples (i.e., $X_p$ is a proper relation), or $X_o$ as the cross-lingual counterpart of $X_s$ for cross-lingual pairs (i.e., $X_p$ is a pair of language tokens).
This task can be formulated into an autoregressive
language modeling training objective:

{
\begin{equation*}
    \mathcal{L}_\text{LM} = -\sum_{x_t\in X_o \cup \{\texttt{[EOS]}\}} \log \mathrm{P}\left(x_t \, | \,x_{<t}\right),
\end{equation*}}%
where $\mathrm{P}\left(x_t \, | \, x_{<t}\right)$ is the conditional probability of generating $x_t$ given previous subtokens.
The probability of generating token $x_t$ is calculated from the hidden state of its previous token $\bm{h}_{t-1}$ in the final layer of Transformer as follows:

{
\begin{equation*}
    \mathrm{P}\left(x_t\,|\,x_{<t}\right)=\text{softmax}(\bm{W} \bm{h}_{t-1}),
\end{equation*}}%
where $\bm{W}$ is a trainable parameter initialized from PLMs for subtoken prediction.
Note that this training objective is applied to both monolingual knowledge triples and cross-lingual links as they can both be encoded in the same $\{s,p,o\}$ format. 




Since models like mBERT or XLM-R rely on masked language modeling which also looks `into the future', subtokens can be leaked by attention.
Therefore, we create adaptations to support causal autoregressive training using attention masks~\cite{yang2019xlnet}, so that the $X_o$ subtokens can only access their previous subtokens.
In particular, in the Transformer blocks, given the query $\bm{Q}$, key $\bm{K}$, and value $\bm{V}$,
we adapt them to a causal LM:

{\small
\begin{equation*}
    \textsc{att}\left(\bm{Q},\bm{K}, \bm{V}\right)=\text{softmax}\left(\frac{\bm{Q}\bm{K}^\intercal}{\sqrt{d}} + \bm{M}\right)\bm{V},
\end{equation*}}%
where $\bm{Q}, \bm{K}, \bm{V}\in \mathbb{R}^{l\times d}$; $l$ is the length of the input sequence, $d$ is the hidden size, $\bm{M}\in \mathbb{R}^{l\times l}$ is an attention mask, which is set as follows:

{
\begin{align*}
    \bm{M}_{ij}=
    \begin{cases}
        0 & x_i \notin X_o \cup \{\texttt{[EOS]}\} \\
        0 & x_i \in X_o \cup \{\texttt{[EOS]}\}, j \le i \\
        -\infty & x_i \in X_o \cup \{\texttt{[EOS]}\}, j > i \\
    \end{cases}
\end{align*}}%

\subsection{Inference}\label{ssec:inference}
Different downstream tasks might require different types of inference: e.g., while link prediction tasks should rely on autoregressive inference, similarity-based tasks such as cross-lingual entity linking rely on similarity-based inference, that is, finding nearest neighbors in the multilingual space. In what follows, we outline both inference types.

\stitle{Autoregressive Inference.}
For link prediction tasks test input is in the format of $\{s, p, ?\}$, where the model is supposed to generate the missing $o$ given $s$ and $p$. 
For such tasks, $o$ comes from a known set of candidate entities $\mathcal{O}$. A simple way to perform inference is to construct candidate tuples $\{s, p, o'\}$ using each $o'\in \mathcal{O}$ and return the one with the minimum LM loss. This straightforward approach requires encoding $|\mathcal{O}|$ sequences. However, as $|O|$ can be large for high-resource languages~(e.g., 2M items for English), this might yield a prohibitively expensive inference procedure. We thus propose to speed up inference by applying and adapting the constrained beam search~\cite{anderson-etal-2017-guided}. In a nutshell, instead of calculating loss on the whole sequence, we generate one subtoken at a time and only keep several most promising sequences in the expansion set for beam search.
The generation process ends when we exceed the maximum length of entities.

More precisely, given $s$ and $p$ (or only $s$ when dealing with cross-lingual links), we concatenate them as the initial sequence $X_0$ and initialize the sequence loss to 0. We then extend the sequence using subtokens from the PLM's vocabulary $\mathcal{V}$. For each subtoken $w_1\in \mathcal{V}$, we create a new sequence $\{X_0, w_1\}$ and add $-\log \mathrm{P}\left(w_1|X_0\right)$ to the sequence loss.
For the next round, we only keep the sequences that can be expanded to an entity in the expansion set, and retain at most $K$ sequences with the smallest sequence loss, where $K$ is a hyperparameter.
This process is repeated until there are no more candidate sequences to be added to the expansion set.
Finally, for any candidate entity $o\in \mathcal{O}$, if it has been generated from a corresponding candidate sequence, we set its loss to the total LM loss~(sum of sequence losses), otherwise we set its loss to $\infty$. Finally, we return the entity with the smallest loss. A more formal description of this procedure is summarized in Alg.~\ref{algo::bs} in the Appendix.

This inference variant only requires encoding at most $L\cdot K$ sequences, where $L$ is the maximum number of subtokens in an entity.
It is much more efficient when $L\cdot K \ll |\mathcal{O}|$, which generally holds for tasks such as link prediction.

\stitle{Similarity-Based Inference.} For some tasks it is crucial to retrieve nearest neighbors (NN) via embedding similarity in the multilingual space. Based on prior findings concerning multilingual PLMs \cite{liu-etal-2021-learning-domain} and our own preliminary experiments, out-of-the-box \modelname produces entity embeddings of insufficient quality. However, we can transform them into entity encoders via a simple and efficient unsupervised Mirror-BERT procedure \citep{liu2021fast}. In short, Mirror-BERT is a contrastive learning method that calibrates PLMs and converts them into strong universal lexical or sentence encoders. The NN search is then performed with the transformed ``Mirror-BERT'' \modelname variant.\footnote{For a fair comparison, we also apply the same transformation on baseline PLMs.}

\section{Experiments and Results}

In this section, we evaluate \modelname in both high-resource and low-resource languages. The focus is on four tasks that are directly or indirectly related to KB construction. 1) Link prediction (LP) is the core task for automatic KB construction since it discovers missing links given incomplete KBs. 2) Knowledge probing from LMs (LM-KP) can also be seen as a type of KB completion task as it performs entity retrieval given a subject entity and a relation. 3) Cross-lingual entity linking (XEL) and 4) Bilingual lexicon induction (BLI) can be very useful for multilingual KB construction as they help to find cross-lingual entity links. 

\subsection{Experimental Setup}
\rtitle{Training Configuration.}
We train our model on knowledge facts for 87 languages which are represented both in DBpedia and in XLM-R (Base). The training set comprises 52M monolingual knowledge triples and 142M cross-lingual links. 


We implement our model using Huggingface's Transformers library~\cite{wolf-etal-2020-transformers},
and primarily follow the optimization hyperparameters of XLM-R.\footnote{In summary: The model is trained for 5 epochs with the Adam optimizer~\cite{Kingma2015AdamAM} using $\beta_1=0.9$, $\beta_2=0.98$ and a batch size of 1,024.
The learning rate is $5\mathrm{e}{-5}$, with a warmup for the first $6\%$ steps followed by a linear learning rate decay to 0.
We use dropout~\cite{srivastava2014dropout} with a rate of 0.1 on all layers and attention weights.
For efficiency, we drop all triples with sequence lengths $\ge 30$, which only constitutes less than $1.3\%$ of all triples.
The full training takes about 5 days with one Nvidia RTX 8000 GPU.} For LP we use the final checkpoint; for LM-LP, results are reported using the checkpoint at 20k steps; for BLI and XEL, the checkpoint at 150k steps is used. We discuss the rationales of checkpoint selection in \Cref{sec:analysis}.

\stitle{Inference Configuration.}
For similarity-based inference, as in previous work \cite{liu2021fast} the Mirror-BERT procedure relies on the 10k most frequent English words for contrastive learning.\footnote{We use English words only for simplicity and direct comparisons. According to \citet{liu2021fast}, Mirror-BERT tuning which uses words from the actual test language pair might yield even better performance. Our training config is identical to the original Mirror-BERT work, except the use of a smaller batch size (128 instead of 200) due to hardware constraints.} For constrained beam search, used with the LP task, we set the hyperparameter $K$ to 50.






\begin{table*}
\centering
\begin{tabular}{llccccccccccccc}
\toprule
& \lpresults{lang.$\rightarrow$}{\en}{\fin}{\de}{\fr}{\tr}{\et}{\hu}{\ita}{\ja}{avg.} \\
\midrule
& \lpresults{\# entities~(K)}{2175}{187}{304}{671}{159}{32}{151}{525}{422}{-} \\
& \lpresults{\# triples~(K)}{7256}{634}{618}{1912}{528}{66}{535}{1543} {1159}{-} \\
\midrule
\multirow{5}{*}{\rotatebox[origin=c]{90}{\emph{Hits@1}}} & \lpresults{TransE}{11.3}{2.4}{4.8}{3.0}{6.1}{2.6}{11.4}{4.1}{1.9}{5.3}  \\
& \lpresults{ComplEx}{15.3}{18.8}{11.6}{16.3}{16.3}{16.3}{15.0}{12.8}{12.7}{15.0} \\
& \lpresults{RotatE}{19.7}{19.8}{17.5}{23.0}{26.2}{21.5}{29.8}{17.3}{15.8}{21.2} \\
& \lpresults{\modelname~(Single)}{25.5}{19.0}{17.8}{23.8}{37.6}{16.1}{32.6}{17.9}{19.7}{23.3} \\
\rowcolor{blue!10}
& \lpresults{\modelname~(All)}{\textbf{27.3}}{\textbf{22.4}}{\textbf{20.8}}{\textbf{25.0}}{\textbf{41.8}}{\textbf{25.8}}{\textbf{35.1}}{\textbf{22.7}}{\textbf{20.6}}{\textbf{26.8}} \\
\midrule
\multirow{5}{*}{\rotatebox[origin=c]{90}{ \emph{Hits@3}}} & \lpresults{TransE}{28.0}{26.0}{24.0}{27.2}{31.0}{20.0}{36.1}{25.0}{20.6}{26.4} \\
& \lpresults{ComplEx}{22.3}{30.1}{20.7}{24.0}{26.9}{24.8}{29.0}{22.2}{22.9}{24.8} \\
& \lpresults{RotatE}{29.6}{\textbf{32.8}}{26.8}{30.1}{37.4}{34.6}{42.6}{28.4}{26.7}{32.1} \\
& \lpresults{\modelname~(Single)}{34.1}{27.6}{24.8}{29.6}{46.1}{25.6}{44.1}{27.7}{\textbf{29.4}}{32.1} \\
\rowcolor{blue!10}
& \lpresults{\modelname~(All)}{\textbf{35.6}}{31.8}{\textbf{29.7}}{\textbf{32.4}}{\textbf{49.8}}{\textbf{36.7}}{\textbf{47.5}}{\textbf{32.2}}{\textbf{29.4}}{\textbf{36.1}} \\
\midrule
\multirow{5}{*}{\rotatebox[origin=c]{90}{ \emph{Hits@10}}} & \lpresults{TransE}{41.4}{\textbf{47.9}}{38.8}{43.5}{50.3}{38.3}{51.0}{42.3}{37.9}{43.5} \\ 
& \lpresults{ComplEx}{32.2}{44.4}{32.7}{35.7}{41.7}{35.6}{45.0}{34.7}{35.5}{37.5} \\
& \lpresults{RotatE}{39.1}{47.7}{40.0}{\textbf{44.9}}{52.3}{46.4}{55.2}{42.2}{\textbf{40.0}}{45.3} \\
& \lpresults{\modelname~(Single)}{42.5}{39.2}{33.3}{37.6}{54.3}{34.8}{55.4}{38.2}{36.7}{41.3} \\
\rowcolor{blue!10}
& \lpresults{\modelname~(All)}{\textbf{44.3}}{44.0}{\textbf{40.1}}{40.3}{\textbf{58.7}}{\textbf{47.5}}{\textbf{56.8}}{\textbf{42.5}}{38.0}{\textbf{45.8}} \\
\bottomrule
\end{tabular}
\caption{Link prediction statistics and results. The languages (see Appendix for the language codes) are ordered based on their proximity to English (e.g., \ita, \de and \fr being close to \en and \hu and \ja are distant to \en; \citealt{chiswick2005linguistic}). \fin, \et, \tr and \hu have less than 1M Wikipedia articles and are relatively low-resource.} 
\label{tab:lp_MRR}
\end{table*}

\begin{table}
\centering
\begin{tabular}{lcccccccccccc}
\toprule
lang.$\rightarrow$ & \te & \lo & \mr & avg. \\
\midrule
XLM-R + Mirror & 2.1 & 4.0 & 0.1 & 2.1 \\
mBERT + Mirror &  3.2 & \textbf{8.0} & 0.1 & 3.8 \\
\rowcolor{blue!10}
\modelname + Mirror & \textbf{13.09} & 7.6 & \textbf{21.0} & \textbf{13.9} \\
\bottomrule
\end{tabular}
\caption{XEL accuracy on the LR-XEL task for low-resource languages.}
\label{tab:lr_xel}
\end{table}

\begin{table*}
\centering
\setlength{\tabcolsep}{5pt}
\begin{tabular}{lcccccccccccccc}
\toprule
lang.$\rightarrow$ & \en & \es & \de & \fin & \ru & \tr & \ko & \zh & \ja & \textsc{th} & avg. \\
\midrule
XLM-R + Mirror & \textbf{75.4} & 34.0 & 13.7 & 4.2 & 7.4 & 19.5 & 1.8 & 1.4 & 2.7 & 3.2 & 16.3 \\
mBERT + Mirror & 73.1 & 40.1 & 16.6 & 4.4 & 5.0 & 22.0 & 1.9 & 1.1 & 2.3 & 2.4 & 16.9 \\
\modelname (Single) + Mirror & \textbf{75.4} & 39.5 & 16.9 & 8.4 & 12.4 & 27.4 & 2.1 & 3.5 & 4.1 & 6.9 & 19.7 \\
\rowcolor{blue!10}
\modelname (All) + Mirror & 71.9 & \textbf{49.2} & \textbf{25.7} & \textbf{15.2} & \textbf{24.5} & \textbf{34.1} & \textbf{9.3} & \textbf{6.9} & \textbf{13.7} & \textbf{14.5} & \textbf{26.5} \\
\bottomrule
\end{tabular}
\caption{XEL Accuracy on XL-BEL.}
\label{tab:xl_bel}
\end{table*}

\begin{table*}[!t] 
\centering
\setlength{\tabcolsep}{3pt}
\begin{tabular}{lcccccccccccccccccc}
\toprule
 \multirow{2}{*}{\shortstack[l]{lang.$\rightarrow$\\ \\ model$\downarrow$}} & \multicolumn{2}{c}{\en-\ita} &  &\multicolumn{2}{c}{\en-\tr} &  &\multicolumn{2}{c}{\en-\ru} & &\multicolumn{2}{c}{\en-\fin} &  &\multicolumn{2}{c}{\fin-\ru} &  &\multicolumn{2}{c}{\fin-\tr} \\
\cmidrule[0.5pt]{2-3}\cmidrule[0.5pt]{5-6}\cmidrule[0.5pt]{8-9}\cmidrule[0.5pt]{11-12}\cmidrule[0.5pt]{14-15}\cmidrule[0.5pt]{17-18}
& Acc & MRR & & Acc & MRR & & Acc & MRR & & Acc & MRR & & Acc & MRR & & Acc & MRR  \\
\midrule
XLM-R + Mirror & \textbf{12.0} & 16.6 && \textbf{6.9} & 8.6 && 2.9 & 5.9 && 5.9 & 7.4 && 2.0 & 3.3 && 5.7 & 7.0  \\
\rowcolor{blue!10}
\modelname + Mirror & 11.5 & \textbf{20.4} && 6.7 & \textbf{11.1} && \textbf{3.7} & \textbf{11.4} && \textbf{6.9} & \textbf{11.5} && \textbf{4.2} & \textbf{9.0} && \textbf{7.7} & \textbf{11.0} \\
\bottomrule
\end{tabular}
\caption{Accuracy and MRR for BLI. mBERT results are omitted since it performs much worse than XLM-R.}
\label{tab:bli}
\end{table*}


\subsection{Link Prediction}

\rtitle{(Short) Task Description.}
Following relevant prior work \cite{bosselut-etal-2019-comet,Yao2019KGBERTBF}, given a subject entity $e_1$ and relation $r$, the aim of the LP task is to determine the object entity $e_2$.

\stitle{Task Setup.}
We evaluate all models on DBpedia. We randomly sample $10\%$ of the monolingual triples as the test set for 9 languages and use remaining data to train the model.\footnote{Following \citet{bordes2013translating}, we use the \emph{filtered} setting, removing corrupted triples appearing in the training or test set.  Moreover, following existing LP tasks \cite{toutanova2015representing,dettmers2018convolutional} we remove redundant triples $(e_1, r_1, e_2)$ from the test set if $(e_2, r_2, e_1)$ appears in the training set.}
The data statistics are reported in \Cref{tab:lp_MRR}. The evaluation metrics are standard  \emph{Hits@1},  \emph{Hits@3}, and  \emph{Hits@10}.\footnote{We do not calculate mean rank and mean reciprocal rank as constrained beam search does not yield full ranked lists.}

\stitle{Models in Comparison.}
We refer to our model as \modelname~(All) and compare it to the following groups of baselines. First, we compare to three representative and widely used KG embedding models\footnote{The KG embedding baselines are implemented based on OpenKE~\cite{han2018openke} and trained using the default hyper-parameters in the library.}:  1) TransE~\cite{bordes2013translating} interprets relations as translations from source to target entities,
2) ComplEx~\cite{trouillon2016complex} uses complex-valued embedding to handle binary relations, while 3) RotatE~\cite{sun2018rotate} interprets relations as rotations from source to target entities in the complex space. 
In fact, RotatE additionally uses a self-adversarial sampling strategy in training, and offers state-of-the-art performance on several KG completion benchmarks \cite{rossi2021knowledge}.
Second, \modelname~(Single) is the ablated monolingual version of \modelname, which uses an identical model structure to \modelname~(All), but is trained only on monolingual knowledge triples of the test language. Training adopts the same strategy from prior work on pretraining monolingual LMs for KG completion \cite{bosselut-etal-2019-comet,Yao2019KGBERTBF}.
We train the \modelname~(Single) for the same number of epochs as \modelname~(All): this means that the embeddings of subtokens in the test language are updated for the same number of times.

\begin{table}
\centering
\setlength{\tabcolsep}{1pt}
\begin{tabular}{lccccccccc}
\toprule
lang.$\rightarrow$ & \en & \ita  & \de & \fr & \fin & \et & \tr &  \hu &  avg. \\
\midrule
XLM-R & 21.0 & 19.3 & 13.9 & 7.6 & 5.6 &  6.1 & 20.5 & 6.1  & 12.5 \\
\rowcolor{blue!10}
\modelname &  \textbf{23.8} & \textbf{21.8} & \textbf{20.7} & \textbf{17.8} & \textbf{16.1} & \textbf{7.4} & \textbf{23.9}  & \textbf{13.1}  & \textbf{18.1} \\
\bottomrule
\end{tabular}
\caption{Accuracy on mLAMA.}
\label{tab:mlama}
\end{table}

\stitle{Results and Discussion.}
The results in \Cref{tab:lp_MRR} show
that the \modelname~(All) achieves the best  \emph{Hits@1} on average, outperforming TransE, ComplEx, and RotatE by $21.5\%$, $11.8\%$, and $5.6\%$, respectively.
It also outperforms the baselines on  \emph{Hits@3} and  \emph{Hits@10}.
Moreover, \modelname~(All) outperforms in almost all languages its monolingual counterpart \modelname~(Single): the average improvements are $>3\%$ across all metrics, demonstrating that the model can effectively leverage complementary knowledge captured and transferred through massive pretraining on multiple languages. 
Interestingly, the advantages of \modelname (both Single and All models) over baselines are not restricted to low resource languages but are observed across the board. This hints that, beyond integrating multilingual knowledge, \modelname is essentially a well-suited framework for KB completion in general.

\subsection{Cross-lingual Entity Linking}

\rtitle{(Short) Task Description.} In XEL\footnote{XEL in our work refers only to entity mention \textit{disambiguation}; it does not cover the mention detection subtask.}, a model is asked to link an entity mention in any language to a corresponding entity in an English KB or in a language-agnostic KB.\footnote{A language-agnostic KB has universal interlingual concepts without being restricted to a specific language.}  XEL can contribute to multilingual KB construction in two ways. First, since XEL links mentions extracted from free text to KBs, it can be leveraged to enrich KBs with textual attributes. 
Second, it also provides a way to disambiguate knowledge with similar surface forms but different grounded contexts.

\stitle{Task Setup.} We evaluate \modelname on two XEL benchmarks: (i) the Low-resource XEL benchmark (LR-XEL; \citealt{zhou2020improving}) and (ii) cross-lingual biomedical entity linking (XL-BEL; \citealt{liu-etal-2021-learning-domain}). LR-XEL covers three low-resource languages \te, \lo, and \mr\footnote{Marathi (\mr, an Indo-Aryan language spoken in Western India, written in Devanagari script), Lao (\lo, a Kra-Dai language written in Lao script) and Telugu (\te, a Dravidian language spoken in southeastern India written in Telugu script).} where the model needs to associate mentions in those languages to the English Wikipedia pages. XL-BEL covers ten typologically diverse languages (see \Cref{tab:xl_bel} for the full list). It requires the model to link an entity mention to entries in UMLS \citep{bodenreider2004unified}, a language-agnostic medical knowledge base.

\stitle{Models in Comparison.} For XEL and all following tasks, we use multilingual MLMs (i.e. mBERT and XLM-R) as our baselines as they are the canonical models frequently used in prior work and have shown promising results in cross-lingual entity-centric tasks \citep{vulic2020probing,liu-etal-2021-learning-domain,kassner-etal-2021-multilingual}. We remind the reader that the `Mirror-BERT' fine-tuning step is always applied, yielding an increase in performance.

\stitle{Results and Discussion.} On LR-XEL, \modelname achieves gains for all three languages over its base model XLM-R. Especially on \mr, where XLM-R and mBERT are almost fully ineffective, \modelname leads to over 20\% of absolute accuracy gain, again showing the effectiveness of incorporating multilingual structural knowledge. On \lo, mBERT is slightly better than \modelname, but \modelname again yields gains over its base model: XLM-R.
On XL-BEL, a large increase is again observed for almost all target languages (see \modelname (All) + Mirror). The only exception is English, where the model performance drops by 3.5\%. This is likely to be a consequence of trading-off some of the extensive English knowledge when learning on multilingual triples. Beyond English, substantial improvements are obtained in other Indo-European languages including Spanish, German and Russian (+10-20\%), stressing the necessity of knowledge injection even for high-resource languages. Like LP, we also experimented with \modelname trained with only monolingual data (see \modelname (Single) + Mirror). Except for English, very large boosts are obtained on all other languages when comparing All and Single models, confirming that multilingual training has provided substantial complementary knowledge.

\subsection{Bilingual Lexicon Induction}

\rtitle{(Short) Task Description.} BLI aims to find a counterpart word or phrase in a target language. Similar to XEL, BLI can also evaluate how well a model can align a cross-lingual (entity) space. 

\stitle{Task Setup.} We adopt the standard supervised embedding alignment setting \cite{glavavs2019properly} of VecMap \cite{artetxe2018robust} with 5k translation pairs reserved for training (i.e., for learning linear alignment maps) and additional 2k pairs for testing. 
The similarity metric is the standard cross-domain similarity local
scaling (CSLS; \citealt{lample2018word}).\footnote{Note that the models are not fine-tuned but only their embeddings are used. Further, note that the word translation pairs in the BLI test sets have $<0.001\%$ overlap with the cross-lingual links used in \modelname training.} We experiment with six language pairs and report accuracy (i.e.,  \emph{Hits@1}) and mean reciprocal rank (MRR).

\stitle{Results and Discussion.} The results are provided in \Cref{tab:bli}. There are accuracy gains observed on 4/6 language pairs, while MRR improves for all pairs. These findings further confirm that \modelname in general learns better entity representations and improved cross-lingual entity space alignments.

\subsection{Prompt-based Knowledge Probing} 
\rtitle{(Short) Task Description.} LM-KP \citep{petroni2019language} queries a PLM with (typically human-designed) prompts/templates such as \textit{Dante was born in \underline{$\ \ \ \  $}.} (the answer should be \textit{Florence}). It can be viewed as a type of KB completion since the queries and answers are converted from/into KB triples: in this case, \{\ent{Dante}, \rel{born-in}, \ent{Florence}\}.

\stitle{Task Setup.} We probe how much knowledge a PLM contains in multiple languages relying on the multilingual LAnguage Model Analysis (mLAMA) benchmark \cite{kassner-etal-2021-multilingual}. To ensure a strictly fair comparison, we only compare XLM-R and \modelname. We exclude multi-token answers as they require multi-token decoding modules, which will be different for causal LMs like \modelname versus MLMs such as XLM-R. For both \modelname and XLM-R, we take the word with highest probability at the \mask ~token as the model's prediction. Punctuation, stop words, and incomplete WordPieces are filtered out from the vocabulary during prediction. \footnote{The exclusion of multi-token answers and also a customised set of non-essential tokens make our results incomparable with the original paper. However, this is a fair probing setup for comparing \modelname and XLM-R since they share the same tokenizer and their prediction candidate spaces will thus be the same.}

\stitle{Results and Discussion.} \Cref{tab:mlama} indicates that \modelname achieves better performance than XLM-R on mLAMA across all languages. We suspect that the benefits of \modelname training are twofold. First, multilingual knowledge is captured in the unified LM representation, which improves LM-KP as a knowledge-intensive task. The effect of this is particularly pronounced on low-resource languages such as \fin, \et and \hu, showing that transferring knowledge from other languages is effective. Second, the \modelname training on knowledge triples is essentially an \textit{adaptive fine-tuning} step \citep{ruder2021lmfine-tuning} that exposes knowledge from the existing PLMs' weights. We will discuss this conjecture, among other analyses, in what follows.

\subsection{Additional Analysis}\label{sec:analysis}
\begin{figure}[!t]
    \centering
    \includegraphics[width=0.95\linewidth]{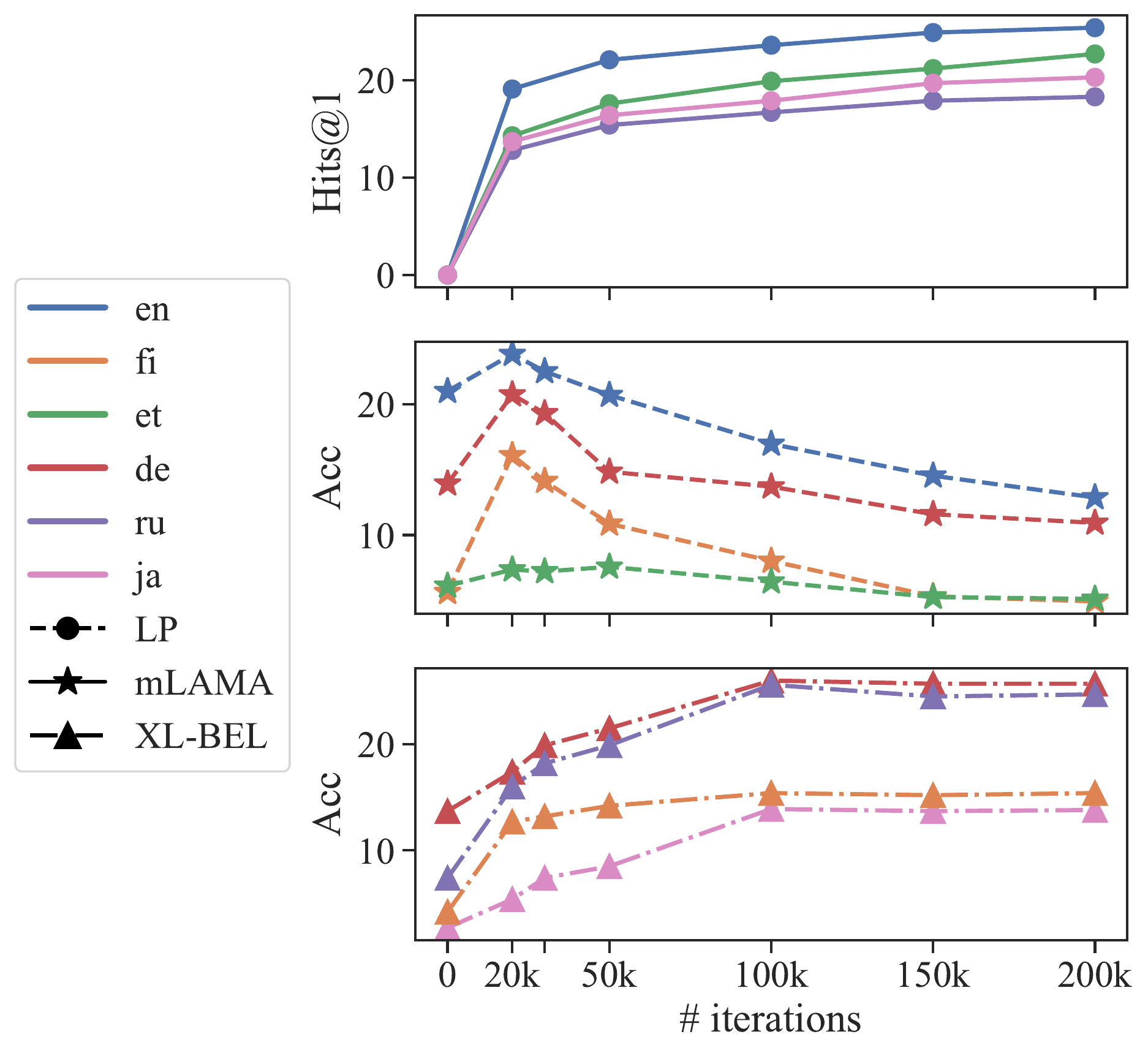}
    \caption{\modelname performance on LP, mLAMA, and XL-BEL over different checkpoints. Results of a sample of languages are shown for clarity.}
    \label{fig:performance_against_checkpoints}
\end{figure}
\rtitle{Inconsistency of the Optimal Checkpoint across Tasks (\Cref{fig:performance_against_checkpoints}).} 
How many steps should we pretrain \modelname on knowledge triples? The plots in \Cref{fig:performance_against_checkpoints} reveal that the trend is different on tasks that require language understanding (mLAMA) versus tasks that require only entity representations (LP and XL-BEL). On mLAMA, \modelname's performance increases initially and outperforms the base model (XLM-R, at step 0). However, after around 20k steps it starts to deteriorate. We speculate that this might occur due to catastrophic forgetting, as mLAMA requires NLU capability to process queries formatted as natural language. Training on knowledge triples may expose the PLMs' capability of generating knowledge at the earlier training stages: this explains the steep increase from 0-20k iterations. However, training on knowledge triples for (too) long degrades the model's language understanding capability. On the other hand, longer training seems almost always beneficial for LP and XL-BEL: these tasks require only high-quality entity embeddings instead of understanding complete sentences. A nuanced difference between LP and XL-BEL is that \modelname's performance on XL-BEL saturates after 100k-150k steps, while on LP the  \emph{Hits@1} score still increases at 200k steps.

\stitle{Link Prediction on Unseen Entities (\Cref{tab:lp_unseen}).}
KG embedding models such as RotatE require that entities in inference must be seen in training. 
However, the \modelname is able to derive (non-random) representations also for unseen entities. We evaluate this ability of \modelname on triples $(s, r, o)$ where the subject entity $s$ or object entity $o$ is unseen during training. The results indicate that \modelname can generalize well also to unseen entities.



\begin{table*}
\centering
\setlength{\tabcolsep}{3.3pt}
\begin{tabular}{lcccccccccccc}
\toprule
lang.$\rightarrow$& \en& \ita& \de& \fr& \fin& \et& \tr& \hu& \ja& avg.\\
\midrule
 \emph{Hits@1}& 17.2& 22.9& 17.0& 16.0& 18.3&31.3& 19.2& 28.5& 12.4& 20.3\\
 \emph{Hits@3}& 24.7& 30.1& 24.0& 22.3& 23.5&37.7& 24.7& 38.5& 19.0& 27.1\\
 \emph{Hits@10}& 31.0& 34.9& 28.9& 27.8& 31.9&42.3& 30.8& 44.2& 23.6& 32.8\\
\bottomrule
\end{tabular}
\caption{LP scores of \modelname~(All) on unseen entities.}
\label{tab:lp_unseen}
\end{table*}




\section{Related Work}

\rtitle{Injecting Structured Knowledge into LMs.} Conceptually, our work is  most related to recent work on knowledge injection into PLMs. 
KnowBERT \citep{peters2019knowledge} connects entities in text and KGs via an entity linker and then re-contextualizes BERT representations conditioned on the KG embeddings. KG-BERT \citep{Yao2019KGBERTBF} trains BERT directly on knowledge triples by 
linearizing their entities and relations into a sequence and predicting plausibility of the sequence. \citet{wang2021structure} improve KG-BERT by splitting a subject-relation-object knowledge triple into a subject-relation pair representation and an object entity representation, then modeling their similarities with a dual/Siamese neural network.\footnote{Other work on knowledge injection such as K-BERT \citep{liu2020k} and ERNIE \citep{zhang-etal-2019-ernie} mainly aims to leverage external knowledge to improve on downstream NLU tasks instead of performing KG completion.} While prior studies have focused on incorporating monolingual (English) structured knowledge into PLMs, our work focuses on connecting knowledge in many languages, allowing knowledge in each language to be transferred and collectively enriched. 

\stitle{Multilingual LMs}  pretrained via MLM, such as mBERT \citep{devlin-etal-2019-bert} and XLM-R \citep{conneau-etal-2020-unsupervised}, cover 100+languages and are the starting point (i.e. initialization) of \modelname.\footnote{We will explore autoregressive multilingual PLMs such as mBART \citep{liu2020multilingual} and mT5 \citep{xue2021mt5} in the future. While they adopt autoregressive training objectives at pretraining, it is non-trivial to extract high-quality embeddings from such encoder-decoder architectures, which is crucial for some tasks in automatic KB completion (e.g. XEL and BLI).}
With the notable exception of \citet{calixto2021wiki} who rely on the prediction of Wikipedia hyperlinks as an auxiliary/intermediate task to improve XLM-R's multilingual representation space for cross-lingual transfer,
there has not been any work on augmenting multilingual PLMs with structured knowledge. Previous work has indicated that off-the-shelf mBERT and XLM-R fail on knowledge-intensive multilingual NLP tasks such as entity linking and KG completion, and especially so for low-resource languages \citep{liu-etal-2021-learning-domain}. These are the crucial challenges addressed in this work. 

\stitle{KB Completion and Construction.} Before PLMs, rule-based systems and multi-staged information extraction pipelines were typically used for automatic KB construction \citep{auer2007dbpedia,fabian2007yago,hoffart2013yago2,dong2014knowledge}.
However, such methods require expensive human effort for rule or feature creation \cite{carlson2010toward,wikidata14}, or they rely on (semi-)structured
corpora with easy-to-consume formats \cite{lehmann2015dbpedia}.
\citet{petroni2019language} showed that modern PLMs such as BERT could also be used as KBs:  querying PLMs with fill-in-the-blank-style queries, a substantial amount of factual knowledge can be extracted. This in turn provides an efficient way to address the challenges of traditional KB methods. \citet{jiang-etal-2020-x} and \citet{kassner-etal-2021-multilingual} extended the idea to extracting knowledge from multilingual PLMs. 

Work in monolingual settings closest to ours is COMET \citep{bosselut-etal-2019-comet}:  \modelname can be seen as an extension of this idea to multilingual and cross-lingual setups. \modelname's crucial property is that it enables knowledge population by transferring complementary structured knowledge across languages. This can substantially enrich (limited) prior knowledge also in monolingual KBs.


In another line of work, multilingual KG embeddings \cite{chen2017multilingual,chen-etal-2021-cross,sun2020alinet,sun-etal-2021-knowing} were developed to support cross-KG knowledge alignment and link prediction. 
Such methods produce a unified embedding space that allows link prediction in a target KG based on the aligned prior knowledge in other KGs \cite{chen-etal-2020-multilingual}.
Research on multilingual KG embeddings has made rapid progress recently, e.g., see the survey of \citet{OpenEA}. However, these methods focus on a closed-world scenario and are unable to leverage open-world knowledge from natural language texts. \modelname combines the best of both worlds and is able to capture and combine knowledge from (multilingual) KGs and multilingual texts.

\section{Conclusion}
We have proposed \modelname, a unified multilingual representation model that can capture, propagate and enrich knowledge in and from multilingual KBs. \modelname is trained via a casual LM objective, utilizing monolingual knowledge triples and cross-lingual links. It embeds knowledge from the KB in different languages into a shared representation space, which benefits transferring complementary knowledge between languages.  We have run comprehensive experiments on 4 tasks relevant to KB construction, and 17 diverse languages, with performance gains that demonstrate the effectiveness and robustness of \modelname for automatic KB construction in multilingual setups. The code and the pretrained models will be available online at: \url{https://github.com/luka-group/prix-lm}.


\section*{Acknowledgement}

We appreciate the reviewers for their insightful comments and suggestions.
Wenxuan Zhou and Muhao Chen are supported by the National Science Foundation of United States Grant IIS 2105329,
and partly by Air Force Research Laboratory under agreement number FA8750-20-2-10002. Fangyu Liu is supported by Grace \& Thomas C.H. Chan Cambridge
Scholarship. Ivan Vuli\'{c} is supported by the ERC PoC Grant MultiConvAI (no. 957356) and a Huawei research
donation to the University of Cambridge.

\bibliography{acl}
\bibliographystyle{acl_natbib}

\clearpage
\appendix

\section{Language Codes}\label{sec:appendix_lang_code}
\begin{table}[!h]
    \centering
    \begin{tabular}{ll}
    \toprule
         \en & English  \\
         \es & Spanish \\
         \ita & Italian \\
         \de & German \\
         \fr & French \\
         \fin & Finnish \\
         \et & Estonian \\
         \hu & Hungarian \\
         \ru & Russian \\
         \tr & Turkish \\
        \ko & Korean \\ 
        \ja & Japanese \\
         \zh & Chinese \\
         \textsc{th} & Thai \\
         \te & Telugu \\
         \lo & Lao \\
         \mr & Marathi \\
    \bottomrule
    \end{tabular}
    \caption{Language abbreviations used in the paper.}
  \label{tab:lang_code}
\end{table}

\section{Constrained Beam Search Algorithm}
The detailed algorithm of constrained beam search is described in~\Cref{algo::bs}.
{
\begin{algorithm}[!t]
    \caption{Constrained Beam Search}\label{algo::bs}
    \setstretch{0.86}
    \normalsize
    \KwInput{Subject entity $s$, relation $p$, set of object entities $\mathcal{O}$, maximum entity length $L$, size of expansion set $K$, PLM vocabulary set $\mathcal{V}$.}
    \KwOutput{Predicted entity.}
    Create the initial sequence $X_0$ by concatenating $s$ and $p$. \\
    Create a set of sequences $\mathcal{X}=\emptyset$. \\
    $\mathcal{X}_0 = \{(X_0, 0)\}$. \\
    \For{$t=1,..., L$}{
        $\mathcal{X}_t = \emptyset$. \\
        \For {$X, l\in \mathcal{X}_{t-1}$}
        {
            \For{$w \in \mathcal{V}$}{
                Add $\left(\{X, w\}, l - \log\mathrm{P}(w_t|X)\right)$ to $\mathcal{X}$ and $\mathcal{X}_t$. \\
            }
        }
        Remove the sequences in $\mathcal{X}_t$ that cannot expand to entities in $\mathcal{O}$. \\
        Keep at most $K$ sequences in $\mathcal{X}_t$ with the smallest loss. \\
    }
    For object entities that appear in $\mathcal{X}$, return the one with the smallest loss.
\end{algorithm}
}

\end{document}